%% file: main.tex
\documentclass[10pt,twocolumn,letterpaper]{article}

\usepackage[pagenumbers]{iccv} 

\input{preamble}

\definecolor{iccvblue}{rgb}{0.21,0.49,0.74}
\usepackage[pagebackref,breaklinks,colorlinks,allcolors=iccvblue]{hyperref}

\def\paperID{00001} 
\def\confName{ICCV}
\def\confYear{2025}

\title{Relative Pose Regression with Pose Auto-Encoders: Enhancing Accuracy and Data Efficiency for Retail Applications}

\author{Yoli Shavit\\
Faculty of Engineering\\ Bar Ilan University \\Ramat-Gan, Israel\\
{\tt\small yoli.shavit@biu.ac.il}
\and
Yosi Keller\\
Faculty of Engineering\\ Bar Ilan University\\ Ramat-Gan, Israel\\
{\tt\small yosi.keller@gmail.com}
}
\begin{document}
\maketitle
\input{sec/0_abstract}    
\input{sec/1_intro}
\input{sec/2_relatedwork}
\input{sec/3_method}

\input{sec/4_results}

\input{sec/5_conclusion}
{
    \small
    \bibliographystyle{ieeenat_fullname}
    \bibliography{main}
}

\end{document}

%% file: preamble.tex
\usepackage[accsupp]{axessibility}  
\usepackage{comment}
\usepackage{gensymb}

%% file: sec/0_abstract.tex
\begin{abstract}
Accurate camera localization is crucial for modern retail environments, enabling enhanced customer experiences, streamlined inventory management, and autonomous operations. While Absolute Pose Regression (APR) from a single image offers a promising solution, approaches that incorporate visual and spatial scene priors tend to achieve higher accuracy. Camera Pose Auto-Encoders (PAEs) have recently been introduced to embed such priors into APR.
In this work, we extend PAEs to the task of Relative Pose Regression (RPR) and propose a novel re-localization scheme that refines APR predictions using PAE-based RPR, without requiring additional storage of images or pose data. 
We first introduce PAE-based RPR and establish its effectiveness by comparing it with image-based RPR models of equivalent architectures. We then demonstrate that our refinement strategy, driven by a PAE-based RPR, enhances APR localization accuracy on indoor benchmarks. Notably,  our method is shown to achieve competitive performance even when trained with only 30\% of the data, substantially reducing the data collection burden for retail deployment. Our code and pre-trained models are available at: \href{https://github.com/yolish/camera-pose-auto-encoders}{https://github.com/yolish/camera-pose-auto-encoders}.
\end{abstract}

%% file: sec/1_intro.tex
\section{Introduction}
\label{sec:intro}
The precise estimation of a camera's position and orientation from visual data is a long-standing challenge in computer vision. Its successful implementation unlocks a multitude of advanced applications within retail, including sophisticated indoor navigation for shoppers, the automated guidance of robotic inventory systems, and the development of augmented reality platforms

Absolute Pose Regression (APR) is a class of methods that directly predict the camera's pose from a single input image \cite{kendall2015posenet}. They are trained with ground-truth pose data to encode images and decode the camera pose parameters. The training is typically performed separately for each scene (e.g. encompassing a single retail space)\cite{kendall2015posenet,shavit2019introduction,kendall2017geometric}, but was recently extended to jointly learn multiple scenes \cite{blanton2020extending,shavit2021learning,shavit2024learning}.

Another approach in camera localization involves regressing the relative motion between a pair of images. If the camera pose of a reference image is known, the relative motion to a new image can be used to determine the new image's pose through straightforward matrix operations. Relative Pose Regression (RPR) methods can offer improved generalization and accuracy compared to APR. However, a drawback of RPR is the requirement for either the original images or their high-dimensional encodings to be available during inference. In contrast, structure-based approaches (e.g., \cite{taira2018inloc,sarlin2019coarse,DSAC,DSAC++,brachmann2023accelerated,wang2024glace}) rely on establishing matches between 2D pixels in an image and 3D scene coordinates. These correspondences are then used to estimate the camera's pose using PnP (Perspective-n-Point) algorithms.

Unlike RPR and structure-based localization methods, APR offers the advantage of not needing to store global or local visual features of the retail environment, nor do they require knowing the intrinsic parameters of the camera capturing the image. This simplifies their deployment in dynamic retail settings. However, this simplicity often comes at the cost of reduced localization accuracy compared to methods that leverage scene-specific visual and spatial features~\cite{pae22}.

Camera Pose Auto-Encoders (PAEs) were recently introduced to augment APR with critical geometric and visual scene information during inference, while maintaining minimal memory and runtime overhead \cite{pae22}. In this work, we present a novel extension of PAEs for RPR and propose a new localization method that leverages this extension. Our method operates in two stages: it first localizes a query image using a pre-trained absolute pose regressor and then refines this initial pose estimate through PAE-based RPR. 

We first demonstrate that PAE-based relative pose regressors achieve localization accuracy comparable to their image-based counterparts, exhibiting consistent performance across different underlying network architectures. We then introduce a Transformer PAE-based RPR architecture specifically designed to implement our proposed localization method. This model is shown to boost the localization accuracy of APR estimates. 

A key advantage of our refinement scheme, compared to previous PAE applications \cite{pae22}, is its ability to enhance both the position and orientation estimates of APR without demanding additional pose storage, test-time optimizations, or image decoding. Furthermore, our method substantially alleviates the data acquisition burden for training absolute pose regressors, achieving competitive accuracy even when trained with only 30\% of a typical general-purpose indoor dataset. We also include  ablation studies to empirically support and validate our design choices.

In summary, our main contributions are as follows:
\begin{itemize}
   \item \textbf{Novel PAE-based Relative Pose Regression}: We introduce a novel extension of PAEs for RPR, demonstrating its efficacy and comparable performance to image-based RPR methods.
   \item \textbf{Two-Stage Localization and Refinement Scheme}: We propose a localization method that utilizes a pre-trained APR model for initial pose estimation, followed by refinement using our novel PAE-based RPR.
   \item \textbf{Reduced Data Requirements and Efficiency}: 
   Our refinement scheme  boosts APR localization accuracy while mitigating the dependency on extensive training data. Our method achieves comparable performance with only 30\% of the data and does not require supplementary pose storage, test-time optimization procedures, or image decoding, making it highly suitable for practical applications.
\end{itemize}

%% file: sec/2_relatedwork.tex
\section{Related Work}
\label{sec:related}
\subsection{Structure-based Methods for Camera Pose Estimation} 
Structure-based pose estimation methods determine a camera's pose by matching detected 2D or 3D feature points to a set of known 3D scene coordinates during inference \cite{taira2018inloc,sarlin2019coarse}, often followed by PnP algorithms. These 3D scene models are typically built using techniques like Structure-from-Motion (SfM)~\cite{sattler2016efficient} or depth sensors \cite{DBLP:conf/cvpr/CavallariGLVST17}. While these approaches offer state-of-the-art localization accuracy and can generalize to scenes not seen during training, they come with significant requirements: they need ground-truth poses, 3D coordinates, local features for reference images, and the intrinsic parameters of both query and reference cameras. Storing image descriptors and 3D feature coordinates for matching can also demand substantial memory. Efforts have been made to reduce this memory footprint, such as quantizing 3D point descriptors \cite{Torii} or using only a subset of 3D points \cite{DBLP:conf/cvpr/CavallariGLVST17}, prioritized by their likelihood of producing valid matches.

An alternative paradigm, Scene Coordinate Regression (SCR), directly estimates 3D scene coordinates from a query image to establish 2D-3D correspondences. While SCR methods achieve state-of-the-art localization accuracy \cite{DSAC,DSAC++,brachmann2023accelerated}, they typically require knowledge of the query camera's intrinsic parameters and exhibit limited generalization to novel environments. Furthermore, scaling SCRs to large-scale environments presents significant challenges, as issues like perceptual aliasing are more pronounced without ground truth 3D supervision \cite{wang2024glace}. To mitigate these limitations, Wang et al. \cite{wang2024glace} suggested to fuse pre-trained local and global features, enabling SCR methods to better cope with the challenges inherent in large-scale environments.
\subsection{Regression Methods for Camera Pose Estimation}
APR methods directly regress camera pose parameters from a query image \cite{kendall2015posenet}. These methods are trained with ground truth pose supervision, encoding the query image into a latent representation from which the pose parameters are then decoded using fully connected layers. Early research explored various encoder and decoder (regressor) architectures \cite{melekhov2017image,naseer2017deep,wu2017delving,shavitferensirpnet,wang2020atloc}, loss formulations \cite{kendall2017geometric,shavitferensirpnet}, and regularization techniques \cite{kendall2016modelling} to enhance performance and mitigate overfitting, aiming to narrow the accuracy gap with structure-based methods.

More recently, research has focused on extending the APR paradigm to jointly learn across multiple scenes \cite{blanton2020extending,shavit2021learning,shavit2024learning}, resulting in improved localization accuracy. Neural Radiance Fields (NeRFs) have also been proposed to augment APR training data \cite{moreau2022lens} and for supervision and test-time optimization \cite{chen2022dfnet}. Furthermore, a recent approach combined SCR and APR learning, integrating an SCR backbone with a transformer-based regressor to achieve state-of-the-art pose estimation accuracy \cite{chen2024map}.

In contrast to APR, RPR methods estimate the motion between a pair of images \cite{balntas2018relocnet,ding2019camnet}. Some research has further leveraged Graph Neural Networks to facilitate information exchange across multiple non-consecutive video frames \cite{9156582,9156582}. During inference, the absolute pose is subsequently derived by estimating the motion from a pose-labeled anchor image. These methods offer enhanced generalization capabilities as they are not intrinsically tied to a specific absolute reference frame. However, their practical application necessitates the availability of a database of pose-labeled anchor images at inference time.

\subsection{Camera Pose Auto-Encoders}
Camera Pose Auto-Encoders (PAEs) \cite{pae22} were recently introduced to incorporate scene priors into APR methods. PAEs achieve comparable or superior performance to their APR teachers by distilling learned representations, enabling various downstream applications, including test-time optimization of position estimate. In the context of RPR, PAEs were proposed as a way for encoding a nearest reference pose to be later decoded into an image and fed into a pre-trained image RPR. 

In this work, we directly extend the PAE paradigm to RPR and propose a localization refinement scheme designed to enhance the performance of APR methods without incurring additional runtime or storage overheads or requiring image decoding and test-time optimization. We demonstrate the effectiveness of our approach using general-purpose indoor benchmarks, which present challenges typical also of retail environments.

%% file: sec/3_method.tex
\section{Method}\label{sec:method}
We present an extension of the PAE paradigm to RPR and introduce a refinement method which leverages this extension to enhance the localization accuracy of APR methods. 

We first establish the necessary technical background by formulating pose auto-encoders (Section \ref{sec:background}). Next, we introduce our extension of PAEs for RPR and describe a transformer-based architecture that implements this paradigm (Section \ref{sec:pae_rpr}). Finally, we present a refinement procedure that updates APR estimates using our PAE-based RPR scheme (Section \ref{sec:LaR-method}).
\subsection{Background}\label{sec:background}
\paragraph{Camera Pose Auto-Encoders.} A camera's pose is defined by its 3D position $\mathbf{x} \in \mathbb{R}^{3}$ and orientation $\mathbf{q} \in {S3}$ (a unit quaternion). A Camera Pose Auto-Encoder (PAE), denoted by $\mathbf{f}$, encodes the \textit{pose} $<\mathbf{x},\mathbf{q}>$ into high-dimensional latent representations
$<\mathbf{\hat{z}_{x}},\mathbf{\hat{z}_{q}}>$. The goal is for these encodings to capture sufficient geometric and visual information, allowing for an absolute pose regressor to accurately decode back the pose parameters. In this work, we follow the work of \cite{pae22} and implement PAEs with sine functions and multi-layer perceptrons (MLPs). 
\paragraph{Training PAEs}
An absolute pose regressor, denoted as $\mathbf{A}$, plays a dual role in training a PAE, $\mathbf{f}$: it acts as both a teacher and a decoder~\cite{pae22}. Given an image $\mathbf{I}$ and its corresponding camera pose $<\mathbf{x},\mathbf{q}>$, the PAE learns by minimizing the following loss:
\begin{equation}
L_{\mathbf{f}}=||\mathbf{z_{x}}-\mathbf{\hat{z}_{x}}||_{2}+||\mathbf{z_{q}}-%
\mathbf{\hat{z}_{q}}||_{2}+L_{\mathbf{p}},  \label{equ:pose encoder loss}
\end{equation}
Here, $\mathbf{z_{x}}$ and $\mathbf{z_{q}}$ are latent representations obtained by encoding $\mathbf{I}$  with $\mathbf{A}$, while $\mathbf{\hat{z}_{x}}$ and $\mathbf{\hat{z}_{q}}$ are the encodings computed by $\mathbf{f}$. The PAE's outputs should enable accurate decoding of the pose parameters when fed into the regressor layers of $\mathbf{A}$. This is achieved by minimizing the camera pose loss, $L_{\mathbf{p}}$, which is commonly employed in training APR methods \cite{kendall2017geometric}:
\begin{equation}
L_{\mathbf{p}}=L_{\mathbf{x}}\exp (-s_{\mathbf{x}})+s_{\mathbf{x}}+L_{%
\mathbf{q}}\exp (-s_{\mathbf{q}})+s_{\mathbf{q}}
\label{equ:learnable_pose_loss}
\end{equation}
In this loss formulation, $L_{\mathbf{x}}$ and $L_{\mathbf{q}}$ represent the position and orientation losses, calculated with respect to a ground truth pose $<\mathbf{x}_{0},\mathbf{q}_{0}>$:
\begin{equation}
L_{\mathbf{x}}=||\mathbf{x}_{0}-\mathbf{x}||_{2}  \label{equ:position loss}
\end{equation}%
and%
\begin{equation}
L_{\mathbf{q}}=||\mathbf{q_{0}}-\frac{\mathbf{q}}{||\mathbf{q}||}||_{2}.
\label{equ:orientation loss}
\end{equation}%
$s_{x}$ and $s_{q}$ are additional learned parameters that encode the uncertainty associated with the position and orientation estimations, respectively \cite{kendall2017geometric}. 

PAEs can be extended to learn from multi-scene APR by additionally encoding the scene index~\cite{pae22}.
\subsection{Relative Pose Regression with PAEs}\label{sec:pae_rpr}
\begin{figure*}[tbh]
\centering
\includegraphics[width=0.8\linewidth]{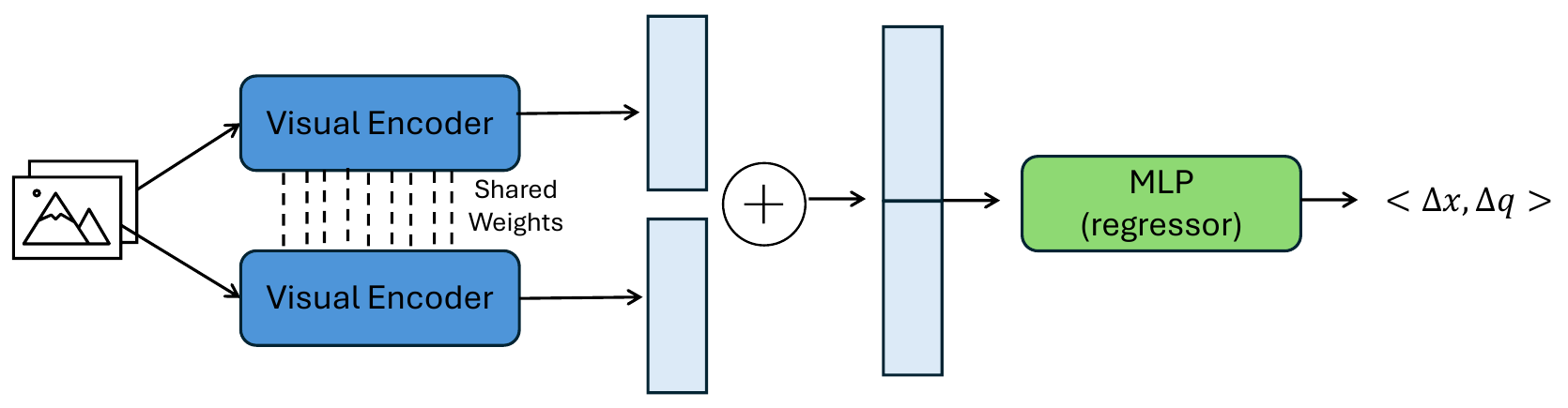}
\caption{Image-based relative pose regression.}
\label{fig:image_based_rpr}
\end{figure*}

\begin{figure*}[tbh]
\centering
\includegraphics[width=0.8\linewidth]{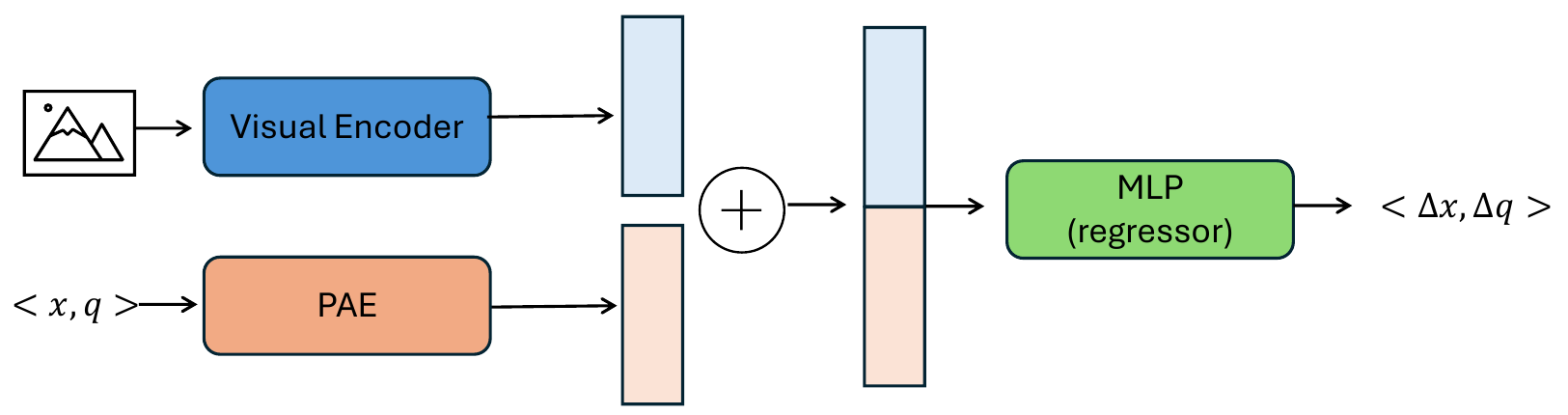}
\caption{PAE-based relative pose regression (our proposed paradigm).}
\label{fig:pae_based_rpr}
\end{figure*}

In this work, we extend the paradigm of PAEs for RPR. 
We first describe image-based RPR and then introduce our proposed PAE-based RPR.
\paragraph{Image-Based RPR.} An image-based relative pose regressor (Fig. \ref{fig:image_based_rpr}) takes a pair of images, a query image and a reference image, as input. A siamese visual encoder (typically, a convolutional neural network) encodes both images into latent tensors. These tensors are then concatenated and fed through one or more MLPs to regress the relative pose.
\paragraph{PAE-based RPR}\label{sec:pae_based_rpr} The PAE-based relative pose regressor (Fig. \ref{fig:pae_based_rpr}) follows a similar process but with a key difference: instead of encoding two images, it encodes a query image and a reference pose. A visual encoder processes the query image, while a PAE encodes the reference pose. The resulting encodings are then used to regress the relative pose.

Both Image-based and PAE-based relative pose regressors are trained by minimizing the camera pose loss (Eq. \ref{equ:learnable_pose_loss}), comparing their estimated relative poses to the ground truth.

\begin{figure*}[tbh]
\centering
\includegraphics[width=0.8\linewidth]{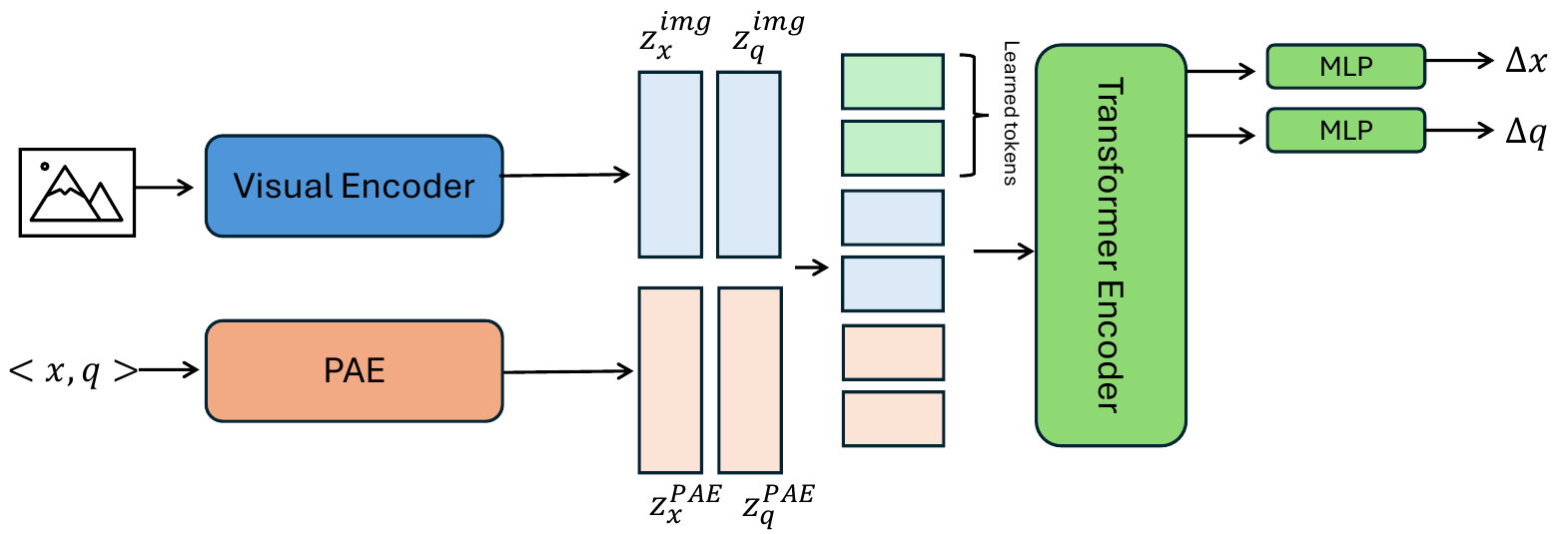}
\caption{Our proposed architecture for implementing PAE-based RPR.}
\label{fig:pae_rpr_transformer}
\end{figure*}

\paragraph{PAE-based RPR with Transformers.} We further introduce a Transformer-based architecture for implementing our proposed PAE-based RPR, as illustrated in Fig. \ref{fig:pae_rpr_transformer}.  This architecture first processes a query image and a reference pose to generate latent position and orientation representations. A convolutional visual encoder is applied to the query image, yielding $\mathbf{z_x^{img}}$, $\mathbf{z_q^{img}}$, while a PAE processes the reference pose to produce $\mathbf{z_x^{pae}}$, $\mathbf{z_q^{pae}}$. All these latent representations have a dimensionality of ${C_d}$ (i.e., are all $\in \mathbb{R}^{C_d}$).

Unlike baseline RPR architectures that use simple concatenation (Fig. \ref{fig:image_based_rpr}), our proposed PAE-based RPR architecture leverages self-attention through a Transformer Encoder \cite{AttentionIsAllYouNeed}. To achieve this, we append two learned tokens, $\mathbf{t_{trans}}\in \mathbb{R}^{C_d}$, 
 for translation and $\mathbf{t_{rot}}\in \mathbb{R}%
^{C_d}$ for rotation, to the sequence of image and pose latent representations. The resulting sequence, of shape $\in \mathbb{R}^{6 \times C_d}$, is then fed into the Transformer Encoder (Fig. \ref{fig:pae_rpr_transformer}).

The Transformer Encoder consists of $L$ identical layers. Each layer $l$, (for $l=1..L$), is composed of a multi-head self-attention module and an MLP module. Following standard practice, Layer Normalization~\cite{ba2016layer} is applied before each module, and residual connections are used to add the input back to the module's output, facilitating stable training. 

We extract the encoder's outputs corresponding to the two learned tokens and then use separate MLPs to regress the translation and rotation parameters.

\subsection{Refining Absolute Pose Regression with PAE-based RPR}
\label{sec:LaR-method} 
Once a PAE is trained, it can encode poses, allowing us to incorporate additional information into pose regression. In this work, we propose to enhance APR by combining it with PAE-based RPR.

A PAE-based relative pose regressor can be used with a trained absolute pose regressor, similar to how image-based RPR localizes a camera. However, PAE-based RPR offers greater flexibility as there is no need to retrieve the closest neighbor from a database. Instead, one can sample or generate a pose near the initial APR estimate, or even use the estimate itself.

\begin{figure*}[tbh]
\centering
\includegraphics[width=0.7\linewidth]{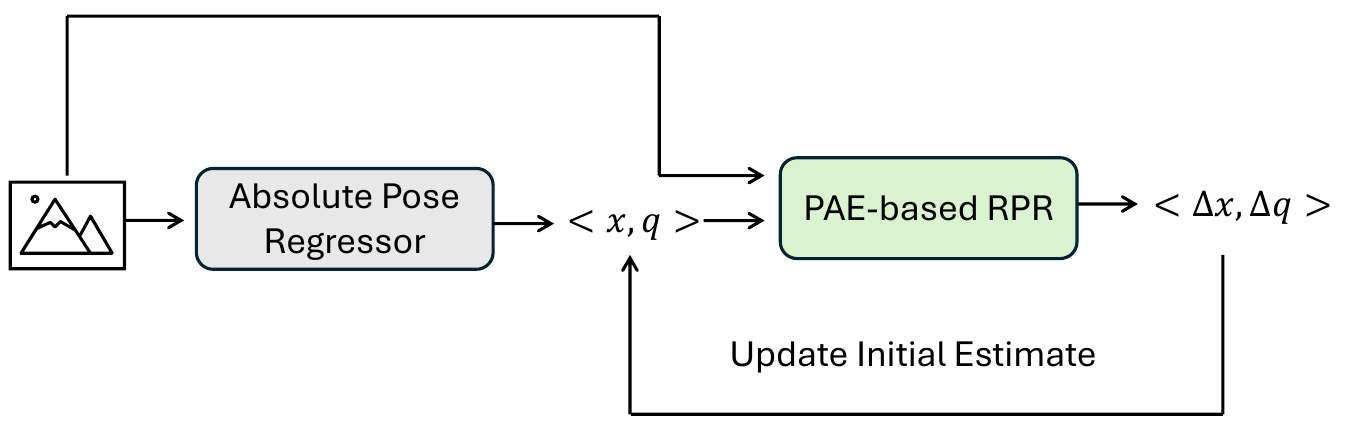}
\caption{Our proposed two-stage localization method, for enhancing APR estimates with PAE-based RPR.}
\label{fig:LaR-method}
\end{figure*}

We propose the following refinement procedure, illustrated in Fig. \ref{fig:LaR-method}:
\begin{enumerate}
    \item A pre-trained absolute pose regressor first estimates the initial pose parameters, $<\mathbf{x}, \mathbf{q}>$, from a query image.
    \item PAE-based RPR then estimates a refinement motion (relative pose), $<\mathbf{\Delta{x}}, \mathbf{\Delta{q}}>$, using the query image and the initial pose estimate.
    \item These refinement motions are used to update the initial pose parameters.
\end{enumerate}
This entire process can be repeated for multiple iterations to further refine the pose estimate.

%% file: sec/4_results.tex
\section{Experimental Results}

\label{sec:Experimental Results}
\subsection{Experimental Setup}
We implement all models and proposed procedures in PyTorch \cite%
{paszke2019pytorch}. Training and inference were performed on an NVIDIA
GeForce GTX 1080 GPU with 8Gb. 

\subsubsection{Datasets}
We evaluated our method using the 7Scenes benchmark \cite{glocker2013real}, a general-purpose dataset for pose regression that features seven compact indoor environments. We selected this dataset because it depicts confined spaces with challenging localization scenarios such as limited visual features and occlusions, which are highly relevant for typical retail environments.
\subsubsection{Implementation and Training Details}
\paragraph{Camera Pose Auto-Encoders.} We implemented and trained PAEs as described in \cite{pae22}.
\paragraph{PAE-based Relative Pose Regression.}
Our refinement process hinges on PAE-based relative RPR. Therefore, we first validated the capacity of relative pose regressors to learn effectively when integrated with PAEs. To achieve this, we implemented and compared the performance of the baseline image-based and PAE-based RPR architectures.

We employed ResNet34, ResNet50, or EfficientNet-B0 as the backbone for image-based RPRs. Two distinct fully connected (FC) layers, each incorporating ReLU non-linearity, were utilized to generate 256-dimensional latent representations for both translation and rotation from each input image. These paired latent vectors were then concatenated and fed into separate multi-layer perceptrons (MLPs) to regress the translation and rotation parameters. Each MLP comprised three fully connected layers with ReLU non-linearity.

The PAE-based RPR model adopted an analogous architecture, with the visual encoder backbone and initial FC layers applied exclusively to the query image. A pre-trained PAE processed the input pose, and two additional, independent FC layers with ReLU non-linearity were employed to produce the corresponding 256-dimensional latent representations for translation and rotation.
For training, we used image pairs from NNnet \cite{nn-net} combined with pairs generated as in CamNet \cite{ding2019camnet}. Both RPR architectures were trained for 20 epochs on the 7Scenes dataset. For PAE-based RPRs, a pre-trained PAE (trained with an MS-Transformer as its teacher) was utilized with its weights frozen. During inference, a pre-trained MS-Transformer \cite{shavit2021learning} was applied, and the neighbor with the closest pose was selected as the reference image/pose for the RPR model.
\paragraph{Refining Absolute Pose Regression with PAE-based RPR}
We implemented our proposed refinement method utilizing the Transformer-based architecture for PAE-based RPR, described in section \ref{sec:pae_based_rpr} (and illustrated in Fig. \ref{fig:pae_rpr_transformer}). We used EfficientNet-B0 as the visual encoder backbone and a pretrained multi-scene PAE.  The Transformer's Encoder comprised two layers with GELU non-linearity and a dropout rate of 0.1. Each layer incorporated a multi-head self-attention module with four heads and a two-layer MLP with a hidden dimension of 2048. The two MLP heads responsible for regressing the translation and rotation vectors, respectively, included a single hidden layer with GELU non-linearity. A pre-trained MS-Transformer was employed to compute the initial pose estimate and the scene index. 

We adopted the same training scheme and dataset used for the baseline PAE-RPRs with a reduced batch size and extended training duration. Specifically, the model was trained for 30 epochs with an initial learning rate of $10^{-4}$ and a batch size of eight. The PAE module was initialized with weights from a pre-trained PAE (trained with an MS-Transformer teacher).

To assess performance on subsets of the dataset, we randomly sampled k\% of images from each scene. This subset was then used to train an MS-Transformer model and a corresponding student PAE according to their respective training protocols. Subsequently, we followed CamNet's procedure to generate training pairs from the given subset and trained our PAE-based RPR as described above. This experiment was conducted for k=70,50,30, with the generation of random subsets repeated across multiple seeds to compute average performance.

\subsection{Comparing PAE-based and Image-based Relative Pose Regression}
Our approach operates on the premise that we can estimate the relative motion of the camera given a query image
and an encoded reference camera pose (without a reference image). To test
this assumption, we compare the performance of image-based and
PAE-based RPR models, with equivalent architectures (i.e., using the same visual encoder and regressor architectures). Table \ref{table:image_vs_pae} shows the
pose errors obtained when localizing the 7Scenes dataset with different
architectures. Image-based and PAE-based RPRs localize images with similar
accuracy. PAE-based RPRs perform favorably in terms of orientation error,
but are slightly less accurate in estimating the camera position.
\begin{table}[tbh]
\caption{{Comparison of image-based and PAE-based RPR models}. We report the median position and orientation errors (meters/degree) for image-based and PAE-based RPR models using various visual encoder backbones. Results are reported on the 7Scenes dataset.}\label%
{table:image_vs_pae}  \centering
\begin{tabular}{ccc}
\toprule \textbf{Visual Encoder} & \textbf{Image-based} & \textbf{PAE-based}
\\
\midrule ResNet34 & 0.22m / 9.53\degree & 0.24m / 9.38\degree \\
ResNet50 & 0.22m / 9.75\degree & 0.23m / 9.45\degree \\
EfficientNet-B0 & 0.21m / 9.36\degree & 0.22m / 8.77\degree \\
\bottomrule &  &
\end{tabular}%
\end{table}
\subsection{Refining Absolute Pose Regression with PAE-based RPR}
\paragraph{Localization Performance.}
We evaluated our proposed refinement scheme (Section \ref{sec:LaR-method}) using a state-of-the-art multi-scene absolute pose regressor (MS-Transformer~\cite{shavit2021learning}). On average, our method improves position error by 5.5\% (from 0.18m to 0.17m) and orientation error by 8.0\%  (from 7.28\degree to 6.69\degree; see Table~\ref{table:subset_training}, first row), while effectively localizing a camera using only the query image as input and handling multiple scenes with a single model. 

\begin{figure*}[th!]
\centering
\includegraphics[scale=0.55]{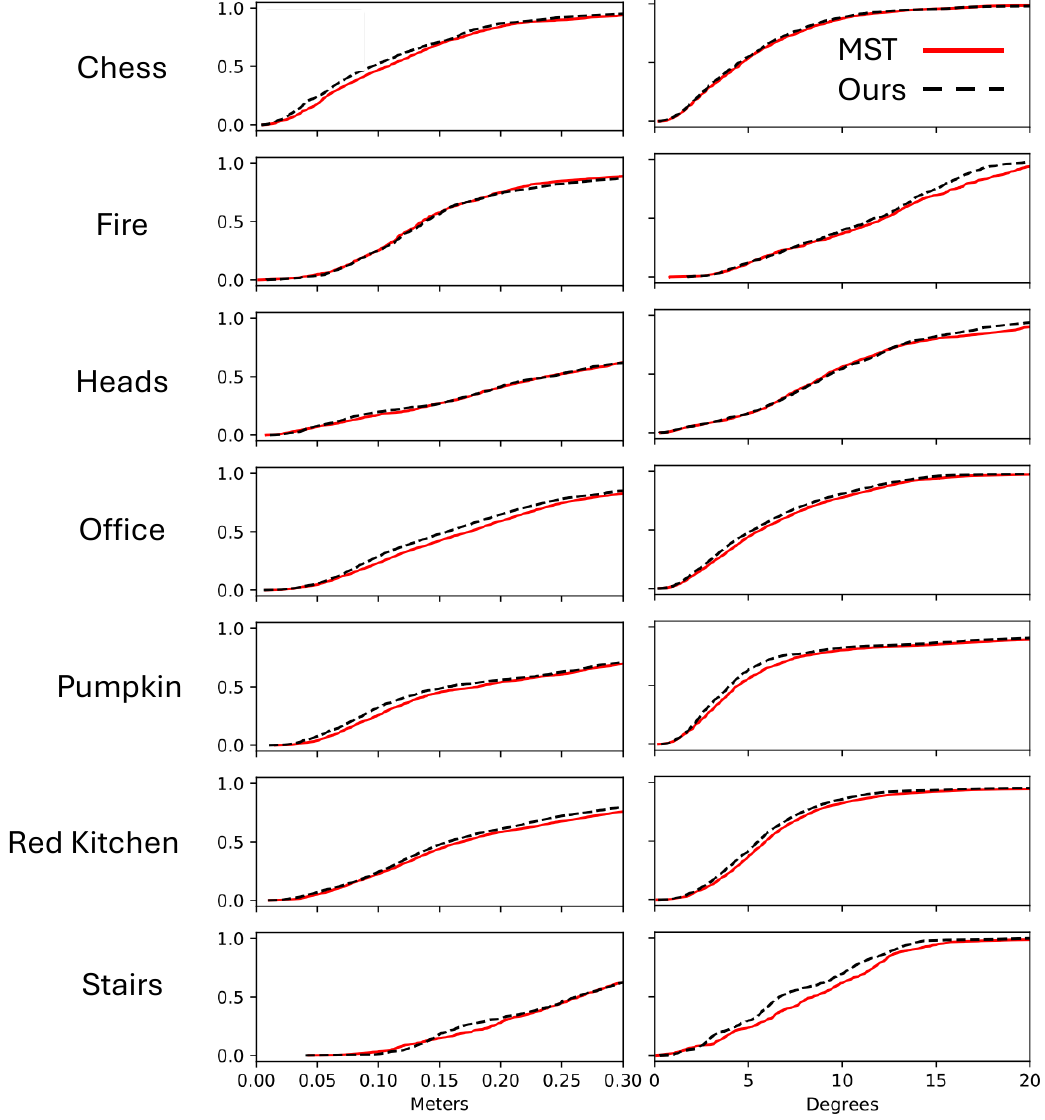}
\caption{Cumulative distributions of position (left column) and orientation
(right column) errors obtained with MS-Transformer (MST, red) and
our proposed localization method (Ours, dashed black), which refines MST estimates with a PAE-based relative pose regressor.
The distributions are truncated at 0.3 meters and $20\degree$, for position
and orientation, respectively.}
\label{fig:cdf}
\end{figure*}

As opposed to previous PAE applications, our PAE-based RPR method enhances both position and orientation accuracy of the APR estimate and does not require additional storage or test-time optimization. Figure \ref{fig:cdf} further illustrates this by comparing the cumulative pose error distributions for the baseline MS-Transformer and our proposed refinement method for each scene. Our approach consistently improves the accuracy of the MS-Transformer, with a particularly noticeable improvement in complex scenes featuring recurring structures, such as the "Stairs" scene.

\paragraph{Training Data Efficiency.}
Relative pose regressors benefit from substantially larger training datasets because they are trained on image pairs derived from the original image set. Since our refinement scheme integrates both absolute and relative pose regression, it potentially requires a smaller training set compared to standalone absolute pose regressors.

Table \ref{table:subset_training} presents the results when training MS-Transformer and our proposed refinement method with varying percentages of the 7Scenes dataset (100\%, 70\%, 50\%, and 30\%). Our method consistently improves the performance of MS-Transformer when trained on the same dataset. Furthermore, we maintain this competitive performance even when trained with only 50\% and even with just 30\%. Notably, our method also exhibits a less significant degradation in performance when less training data is available.
\begin{table}[!th]
\caption{{Data Efficiency of Our PAE-Based RPR Refinement}. We compare MS-Transformer\protect\cite{shavit2021learning} and
MS-Transformer with our PAE-based RPR refinment method, when trained on 100\%, 70\%, 50\% and 30\% of the images in the
training set. Our method demonstrates improved data efficiency by achieving superior or comparable performance even with reduced training data. Results are reported as the mean of median position/orientation errors on the 7Scenes dataset.}\label%
{table:subset_training}  \setlength{\tabcolsep}{2pt} \centering{\
\begin{tabular}{ccc}
\toprule \textbf{\ \% of } & \textbf{MS-Transformer} & \textbf{MS-Transformer with} \\
\textbf{Training Set} &  &  \textbf{PAE-based RPR}\\
\midrule 100\% & 0.18m / 7.28\degree & 0.17m / 6.69\degree \\
70\% & 0.18m / 7.56\degree &0.18m / 7.28\degree \\
50\% & 0.18m / 7.73\degree & 0.18m / 7.32\degree \\
30\% & 0.20m / 8.16\degree & 0.19m / 7.62\degree \\
\bottomrule &  &
\end{tabular}
}
\end{table}
\begin{table}[tbh!]
\caption{Ablations of the number of iterations ($i$) applied when using our
refinement scheme and when using the training pose that is closest to the original pose
estimated (instead of the estimate itself). In the latter case,
we additionally need to store a database of reference poses. We report the
mean of median position/orientation errors in meters/degree, obtained when
localizing the 7Scenes dataset. The best result is highlighted in bold.}\label{ablations:iterations} %
\setlength{\tabcolsep}{2pt} \centering{\
\begin{tabular}{ccc}
\toprule \textbf{Reference Pose} & \textbf{Database} & \textbf{Pose Error}
\\
\midrule Nearest pose & Y & 0.18m / 7.03\degree \\
$i=1$ & N & 0.17m / 6.71\degree \\
$i=2$ & N & 0.17m / 6.70\degree \\
$i=3$ & N & \textbf{0.17m} / \textbf{6.69\degree} \\
\bottomrule &  &
\end{tabular}
}
\end{table}
\begin{table}[tbh!]
\caption{Ablations of the number of Transformer Encoder layers in our PAE-based RPR architecture. We report the mean of the median position/orientation
errors in meters/degree, obtained when localizing the 7Scenes dataset with our proposed PAE-based RPR refinement method. The best result is highlighted in bold. }\label%
{ablation:layers}  \setlength{\tabcolsep}{2pt} \centering{\
\begin{tabular}{cc}
\toprule \textbf{\#Layers} & \textbf{Pose Error} \\
\midrule 2 & \textbf{0.17m} / \textbf{6.71\degree} \\
4 & 0.17m / 6.86\degree \\
6 & 0.18m / 6.75\degree \\
8 & 0.18m / 7.10\degree \\
\bottomrule &
\end{tabular}
}
\end{table}

\subsection{Runtime and Memory Requirements} Our refinement method introduces an additional computational cost compared to standard APR: it requires storing the PAE-based RPR model and executing its forward pass for one or more iterations. However, each iteration of our model takes approximately 37 milliseconds, which is comparable to the runtime of an APR, and the model size is 86 MB, which is a relatively minor requirement.

A key advantage of our refinement method over image-based RPR techniques is its constant memory footprint. Our method eliminates the need for image retrieval or storing RPR encodings of database images, which typically consume several gigabytes of storage space.

\subsection{Ablation Study} 
We conducted ablation studies to evaluate the design choices for both our refinement approach and the proposed PAE-based RPR model. While our primary scheme applies refinement only once, we also investigated its performance with iterative applications.
\paragraph{Iterative Refinement.}
Table \ref{ablations:iterations} shows the pose estimation error when applying the pose refinement process for one, two, and three iterations. When refining with multiple iterations, the updated pose from the previous step serves as the initial estimate for the subsequent refinement. Our method improves the localization accuracy of the initial estimate already after a single iteration and additional improvement is gained with further iterations. We note that while the best estimation is obtained with three iterations, the improvement over one and two iterations is minor. Interestingly, using the closest pose from the training database instead of the APR estimate degrades performance, though still yielding improved localization accuracy compared to MS-Transformer.
\paragraph{PAE-Based RPR Architecture.}
We also evaluated the architectural choices for our PAE-based RPR Transformer model. Table \ref{ablation:layers} presents the pose error when using a Transformer's Encoder with two, four, six, and eight layers. The optimal performance was achieved with a two-layer encoder, which we selected for our implementation. We attribute the inferior performance of larger variants to overfitting. Notably, a competitive performance is maintained across all four architectural variants.

\subsection{Limitations and Future Work}
As PAEs distill knowledge from their absolute pose regressor teachers, they inherit a limitation regarding generalization to scenes not encountered during training. This poses a disadvantage compared to image-based relative pose regressors, which, while still experiencing performance degradation in novel environments, are not inherently restricted to previously seen scenes.

Furthermore, in this work, we exclusively evaluated our method using a single reference image and pose estimate. In future work, we plan to expand this paradigm to incorporate multiple reference images and to extend its applicability beyond compact scenes to large-scale indoor environments.

%% file: sec/5_conclusion.tex
\section{Conclusion}\label{sec:conclusion}
We introduced a novel two-stage camera localization method that enhances the accuracy of APR methods, while addressing their inherent limitations. By extending PAEs to facilitate RPR, we refine initial APR estimates with critical geometric and visual scene information, while maintaining minimal memory and runtime overhead during inference.

Our experiments demonstrate that out PAE-based RPR paradigm achieves localization accuracy comparable to traditional image-based RPR methods. Furthermore, the proposed Transformer PAE-based RPR architecture effectively boosts the localization accuracy of APR estimates, improving both position and orientation. A key advantage of our refinement scheme is its efficiency; it eliminates the need for additional pose storage, test-time optimizations, or complex image decoding.

Importantly, our method substantially alleviates the data acquisition burden associated with training absolute pose regressors, showing competitive performance even when trained with only 30\% of a typical indoor dataset. This makes our approach particularly well-suited for dynamic retail environments where data collection can be challenging. By combining the benefits of APR's simplicity with the refinement capabilities of PAE-based RPR, we offer a robust and efficient solution for camera pose estimation in various retail applications.